%% file: its_ap_2018.tex
\documentclass{llncs}

\usepackage[cmex10]{amsmath}                    
\usepackage{array}                              
\usepackage[noadjust]{cite}                     
\usepackage{subfig}                             
\usepackage{xspace}                             
\usepackage[USenglish]{babel}					
\usepackage{pgfplots}                           
\usepackage{calc}                               
\usepackage[breaklinks=true,hidelinks,          
            bookmarksnumbered=true]{hyperref}   
\usepackage[capitalize]{cleveref}
\crefname{figure}{Figure}{Figures}

\newlength\figurewidth
\newlength\figureheight
\pgfplotsset{every axis/.append style={
		     scaled y ticks=false,
		     scaled x ticks=false,
		     y tick label style={/pgf/number format/fixed},
		     x tick label style={/pgf/number format/fixed},
		     legend style={font=\small}},
	         compat=1.9}                        
\usetikzlibrary{arrows.meta}                    
\usetikzlibrary{external}                       

\newcommand*{\etal}{et al.\@\xspace}                            
\newcommand*{\lsep}{\,---\,}                                    

\graphicspath{{figures/}}                           
\newlength\nwidth                                   
\newlength\nheight                                  
\newlength\nsep                                     
\definecolor{TNOred}{RGB}{204,0,0}                  
\definecolor{TNOlightblue}{RGB}{153,204,255}        
\definecolor{TNOlightgray}{RGB}{222,222,231}        
\tikzstyle{block}=[%
	node distance=\nwidth+\nsep,
	text width=0.88\nwidth,
	minimum height=\nheight,
	minimum width=\nwidth,
	align=center,
	font=\footnotesize\sffamily,
	fill=TNOlightblue,
	rounded corners=0.4em,
	anchor=south west]               
\tikzstyle{halfblock}=[node distance=\nwidth+\nsep,
	text width=0.88\nwidth,
	minimum height=0.5\nheight-0.5\nsep,
	minimum width=\nwidth,
	align=center,
	font=\footnotesize\sffamily,
	fill=TNOlightblue,
	rounded corners=0.4em,
	anchor=south west]               
\tikzstyle{doublehalfblock}=[node distance=2\nwidth+2\nsep,
	text width=1.88\nwidth,
	minimum height=0.5\nheight-0.5\nsep,
	minimum width=2\nwidth+\nsep,
	align=center,
	font=\footnotesize\sffamily,
	fill=TNOlightblue,
	rounded corners=0.4em,
	anchor=south west]               

\begin{document}

\selectlanguage{USenglish}
\setlength\nwidth{5.6em}    
\setlength\nheight{7em}     
\setlength\nsep{0.2em}      

\title{Scenario-Based Safety Assessment Framework for Automated Vehicles}
\author{%
	Jeroen Ploeg\inst{1} \and 
	Erwin de Gelder\inst{1}\thanks{Corresponding author, e-mail: \email{erwin.degelder@tno.nl}.} \and 
	Martin Slav{\'i}k\inst{2} \and 
	Eley Querner\inst{2} \and \\
	Thomas Webster\inst{3} \and 
	Niels de Boer\inst{3}
}
\institute{%
	TNO Singapore, 2 Science Park Drive \#02-07 Tower A, Singapore 118222 \and 
	T\"UV S\"UD Asia Pacific Pte Ltd, Singapore \and 
	Nanyang Technological University, Singapore
}

\maketitle

\input{sections/abstract}
\input{sections/introduction}
\input{sections/automotive_safety}
\input{sections/assessment_pipeline}
\input{sections/scenario_generation}
\input{sections/conclusion}
\input{sections/acknowledgement}

\bibliography{@pubDatabase}
	
\end{document}

%% file: sections/abstract.tex
\begin{abstract}
    Automated vehicles (AVs) are expected to increase traffic safety and traffic efficiency, among others by enabling flexible mobility-on-demand systems. 
    This is particularly important in Singapore, being one of the world's most densely populated countries, which is why the Singaporean authorities are currently actively facilitating the deployment of AVs. 
    As a consequence, however, the need arises for a formal AV road approval procedure. 
    To this end, a safety assessment framework is proposed, which combines aspects of the standardized functional safety design methodology with a traffic scenario-based approach. 
    The latter involves using driving data to extract AV-relevant traffic scenarios. 
    The underlying approach is based on decomposition of scenarios into elementary events, subsequent scenario parametrization, and sampling of the estimated probability density functions of the scenario parameters to create test scenarios.
    The resulting test scenarios are subsequently employed for virtual testing in a simulation environment and physical testing on a proving ground and in real life. 
    As a result, the proposed assessment pipeline thus provides statistically relevant and quantitative measures for the AV performance in a relatively short time frame due to the simulation-based approach. 
    Ultimately, the proposed methodology provides authorities with a formal road approval procedure for AVs. In particular, the proposed methodology will support the Singaporean Land Transport Authority for road approval of AVs.
\end{abstract}

%% file: sections/introduction.tex
\section{Introduction}

The development of automated vehicles (AVs) has made significant progress. It is expected that before 2020, automated and autonomous vehicles will be introduced in controlled environments, whereas autonomous vehicles will be mainstream by 2040 \cite{Madni2018} or earlier \cite{Bimbraw2015}. Especially in densely populated cities such as Singapore, there is a need for automated vehicles to increase traffic safety and traffic efficiency by enabling flexible, automated, mobility-on-demand systems \cite{Spieser2014}, scheduled services for public transport needs, and automated freight and service vehicles to support 24 hours operations and labor shortage needs.

An important aspect in the development of autonomous vehicles (AVs) is the safety assessment of the AVs \cite{Bengler2014, Stellet2015, Helmer2017, Putz2017, Wachenfeld2016}. For legal and public acceptance of AVs, a clear definition of system performance is important, as are quantitative measures for the system quality. The more traditional methods \cite{ISO26262, Response2006}, used for evaluation of driver assistance systems, are no longer sufficient for assessment of the safety of higher level AVs, as it is not feasible to complete the quantity of testing required by these methodologies \cite{Wachenfeld2016}. Therefore, the development of assessment methods is important to not delay the deployment of AVs \cite{Bengler2014}.

We propose a methodology for the safety assessment of AVs that includes functional safety \cite{ISO26262}, safety of the intended functionality \cite{ISO21448}, and behavioral safety \cite{Waymo2017}, ultimately resulting in a qualitative and quantitative assessment of the safety of the AV. This is achieved by adopting a traffic scenario-based assessment \cite{Stellet2015, Putz2017, Gelder2017}.
The traffic scenarios are obtained through analysis of real-world driving data, after which the scenarios are verified regarding their relevance for the system under test. The scenarios are used for virtual and physical safety validation of the system under test, ultimately resulting in a safety report providing road approval authorities with quantitative measures regarding the safe performance of the system under test.

The proposed assessment methodology provides an efficient procedure for road approval for AVs, as the assessment methodology allows for three go/no-go decisions, based on the design review, the virtual safety validation, and, finally, a safety report based on all tests. Furthermore, using test scenarios derived from real-world driving data, the methodology provides quantitative measures regarding the safe performance of the AV in real-world traffic.

The proposed methodology provides authorities with a formal road approval procedure for AVs. In particular, the proposed methodology will be used by CETRAN\footnote{Centre of Excellence for Testing and Research of Autonomous Vehicles at NTU, Singapore.} to support the Land Transport Authority from Singapore for road approval of AVs.

The structure of the paper is as follows. A background of relevant automotive safety notions are detailed in \cref{sec:automotiveSafety}. Next, in \cref{sec:theSafetyAssessmentPipeline}, the proposed safety assessment pipeline is described. \Cref{sec:scenarioGeneration} provides more details on the method of scenario generation after which \cref{sec:conclusion} summarizes the main conclusions.

%% file: sections/automotive_safety.tex
\section{Automotive safety}
\label{sec:automotiveSafety}

The field of automotive safety is very broad, characterized by many types of safety and definitions of the various aspects of safety. Nevertheless, on a high abstraction level, two types of safety are commonly distinguished, being \emph{functional safety} and \emph{safety of the intended functionality} (SOTIF). A third type of safety, being \emph{behavioral safety} can also be identified, relating to how the AV is programmed to behave. 

\begin{figure}[tb]
    \centering
    \input{figures/operational_safety.tikz}
    \caption{Operational safety components}
    \label{fig:operationalSafety}
\end{figure}
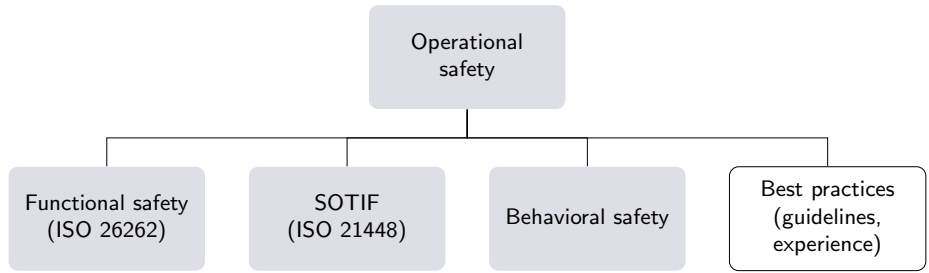

Functional safety, SOTIF, and behavioral safety can be jointly referred to as \emph{operational safety}, as illustrated in \cref{fig:operationalSafety}. This figure also mentions \emph{best practices}, which refers to the use of experience and (non-standard) guidelines, commonly included due to limitations of the ISO\,26262 standard. SOTIF intends to address these limitations, but given the fact that the SOTIF standard is still under development, it is expected that best practices will remain useful, for the time being.

The three types of safety as introduced here, are briefly explained in the next sections.

\subsection{Functional safety}
\label{functionalSafety}

Functional safety is defined in IEC\,61508 as the absence of unacceptable risk due to hazards caused by malfunctioning behavior of systems. ISO/DIS\,26262 \cite{ISO26262}, commonly referred to as ISO\,26262, is an adaptation of the IEC\,61508 for the automotive industry. It applies to safety-related electrical/electronic systems, specifically for road vehicles under 3500\,kg.

The current ISO\,26262 standard was introduced in 2011 and does not include automated vehicles. Advanced driver assistance systems (ADAS) will be included in the next release of ISO\,26262, scheduled for 2018. In ISO\,26262, the notion of functional safety has been defined as follows.

\begin{definition}[Functional safety \cite{ISO26262,Dubey2017}]
    Functional safety is the absence of unreasonable risk due to hazards caused by malfunctioning behavior of electrical/electronics systems.
\end{definition}

\begin{figure}[tb]
    \centering
    \includegraphics[width=\textwidth]{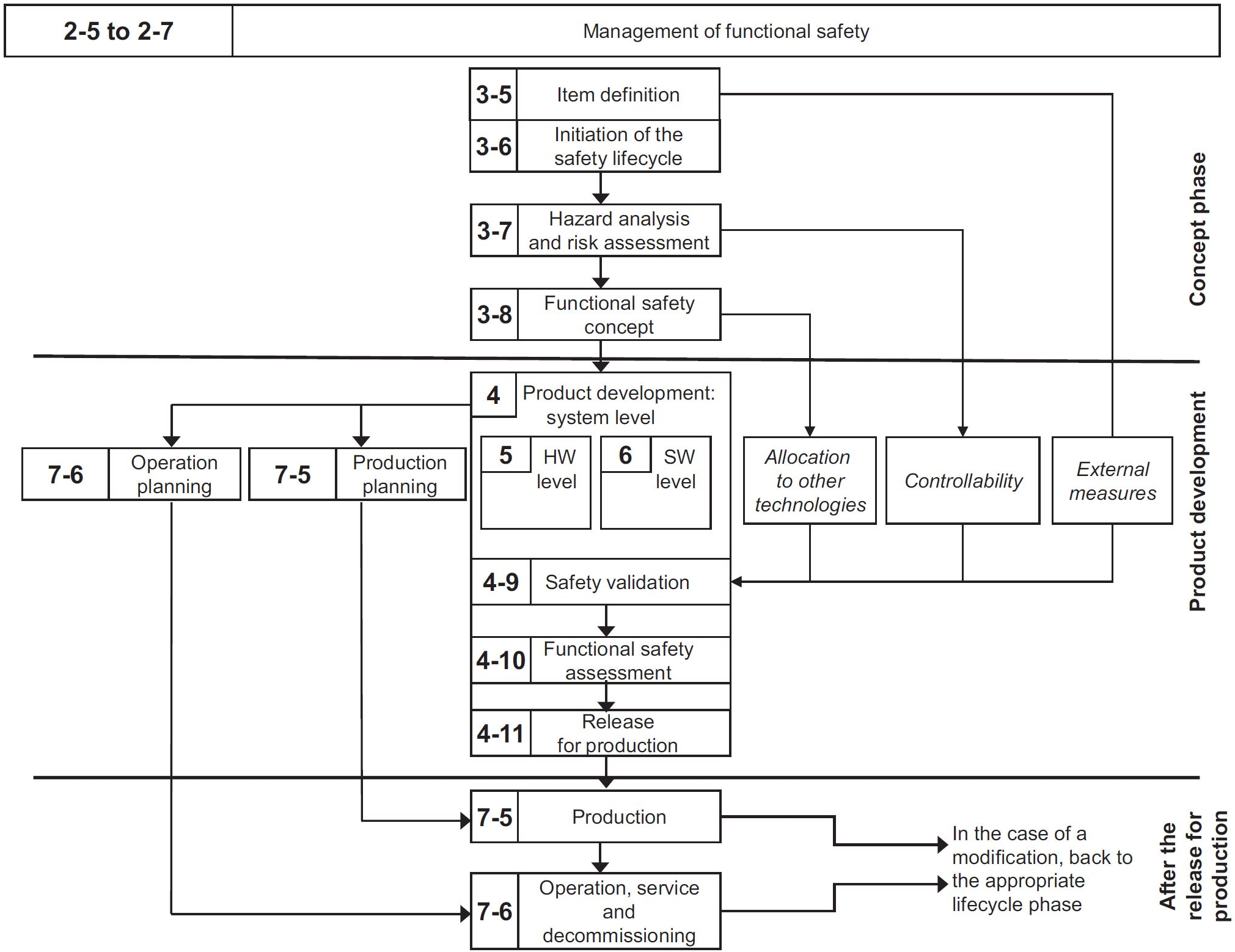}
    \caption{The ISO\,26262 safety lifecycle (extracted from \cite{ISO26262}).}
    \label{fig:ISO26262}
\end{figure}

According to this definition, functional safety is `looking inwards', focusing on malfunctioning components of the vehicle. The process that must be followed according to the ISO\,26262 to arrive at a functionally safe system design, the so-called `safety lifecycle', is shown in \cref{fig:ISO26262}. In the scope of AV safety assessment, the validation processes included in this lifecycle are particularly relevant; These are briefly summarized as follows.

\begin{itemize}
    \item[3-7] \emph{Hazard analysis and risk assessment (HARA)}\lsep The HARA is a method to identify and categorize hazardous events, to specify safety goals, and to assign automotive safety integrity levels (ASILs) to system components, related to the prevention or mitigation of the associated hazards in order to avoid unreasonable risk.
    \item[3-8] \emph{Functional safety concept}\lsep This involves specification of functions and their interactions which are necessary to achieve the desired safe behavior. As such, it is a step in the design, rather than an assessment-oriented activity. However, \emph{review} of the functional safety concept should certainly be part of the AV safety assessment framework. It is noted that the \emph{technical safety concept} is omitted here, since that is typically part of the product development.
    \item[4-9] \emph{Safety validation}\lsep The safety validation checks, on a vehicle level, whether the safety goals are fulfilled. This is done through virtual and physical testing, representing the `intended use cases', i.e., the traffic scenarios as they occur in the envisioned operational environment of the AV.
\end{itemize}

The key notion in the above list is \emph{safety validation} which, in fact, is the ultimate objective of the proposed assessment framework presented in the next section. In ISO\,26262, this notion is defined as follows.

\begin{definition}[Safety validation \cite{ISO26262}]
    Safety validation is the assurance, based on examination and tests, that the safety goals are sufficient and have been achieved.
\end{definition}

\begin{figure}[tb]
    \centering
    \includegraphics[width=\textwidth]{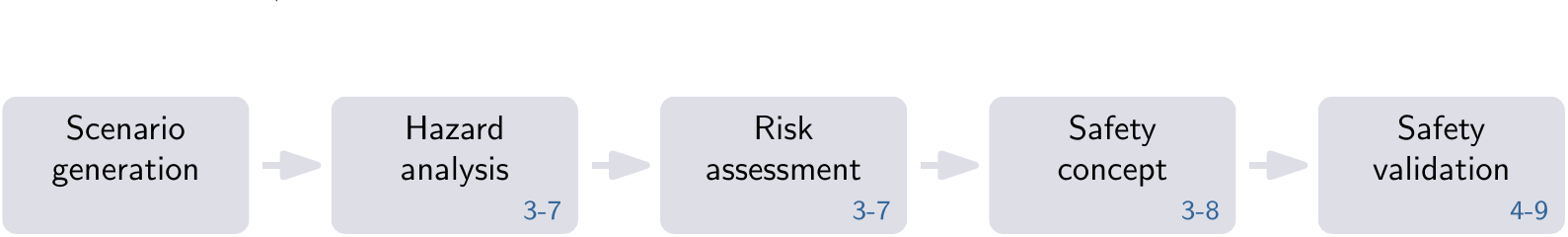}
    \caption{Functional safety assessment steps, inspired by the ISO\,26262.}
    \label{fig:functionalSafetyAssessment}
\end{figure}

When adopting the selected topics of the lifecycle, a first step is made towards a process for safety assessment of AVs from the perspective of functional safety; This is visualized in \cref{fig:functionalSafetyAssessment}, where the scenario generation component, which is not included in the original ISO process, aims to generate relevant traffic scenarios (either or not as part of overarching `use cases').

In summary, it can be concluded that the ISO\,26262 standard is oriented towards development of a certain automation functionality, as part of which verification and validation processes are implemented. The objective of the assessment framework, however, is not to design a safe AV but to assess the safety of a given design. Consequently, the process from \cref{fig:functionalSafetyAssessment} needs to be further adapted, both in terminology and in contents, to better fit this objective.

\subsection{Safety of the intended functionality (SOTIF)}

As mentioned earlier, functional safety is `inward looking', in the sense that it focuses on malfunctioning components of the vehicle's automation system. For AVs, however, it is also important to `look outwards', i.e., to include inherent sensor and perception system limitations in the safety assessment framework and to take decision-making capabilities of the AV into account. These aspects of AV safety are covered by the notion of \emph{safety of the intended functionality} (SOTIF).

SOTIF relates to hazardous behavior which is not caused by a malfunction. In particular for AVs, SOTIF is often related to environmental perception \cite{Fischer2017} since environmental sensors typically have inherent limitations, which may cause the AV to behave unsafely. Examples of such limitations are the occurrence of so-called `ghosts' in radar measurements, or the blinding effect of direct sunlight in case of camera systems. This type of safety is described in the standard ISO/WD\,PAS\,21448, in short referred to as ISO\,21448, which is currently being drafted \cite{ISO21448}. This draft standard defines SOTIF in a very broad sense as ``the absence of unreasonable risk of the intended functions''. In the scope of the assessment framework, a more precise definition will be adopted, as proposed below. 

\begin{definition}[Safety of the intended functionality]
    Safety of the intended functionality (SOTIF) is the ability of the automation system to correctly comprehend the environmental situation and behave safely, by ensuring that the systems and components are operating within their design boundaries and, if they are not, by activating appropriate countermeasures.
\end{definition}

\subsection{Behavioral safety}

Next to SOTIF, in a recent report by Waymo \cite{Waymo2017}, the notion of \emph{behavioral safety} is introduced as ``An aspect of system safety that focuses on how a system should behave normally in its environment to avoid hazards and reduce the risk of mishaps: for instance, detect objects and respond in a safe way (slow down, stop, turn, lane change, etc.).'' This type of safety thus addresses the AV behavior as intentionally programmed by the developer, with all components and subsystems of the vehicle's automation system operating as intended. A relevant question in this context is, for instance, whether the AV follows the traffic rules. Since it is questionable if this type of safety will be part of the final version of ISO\,21448, behavioral safety is regarded as a third type of safety.

The UN-ECE Automatically Commanded Steering Function group is drafting amendments to UN-ECE Regulation 79 which describes the expected behavior of a level 3 automated driving system \cite{SAEJ3016}. However, the scope is limited to well-structured and controlled highways where there are no pedestrians/bicycles, at least two lanes in the direction of travel, and a physical separation between contraflow traffic. The authors are not aware of existing art describing the behavioral safety requirements for a level 4 automated driving system operating in complex urban environments, hence this study aims to provide some definition to the behavioral characteristics required for a vehicle using the context of Singapore’s road traffic rules and traffic environment as the case study. The proposed assessment methodology makes use of traffic scenarios to assess the behavioral safety of the AV. The safety assessment pipeline, which is described in \cref{sec:theSafetyAssessmentPipeline}, sets out how the scenarios are defined from a design review exercise and from scenario generation using acquisition of real-world driving data. Next, the AV's behavioral performance is assessed through virtual and physical safety validation using these scenarios.

%% file: figures/operational_safety.tikz
\tikzstyle{block}=[minimum width=8.5em, minimum height=4.5em, align=center, font=\footnotesize\sffamily, rounded corners=0.4em, fill=TNOlightgray]%
\resizebox{\textwidth}{!}{%
\begin{tikzpicture}
	\node[block](operational){Operational\\safety};
    \node[coordinate, below of=operational, node distance=7em](below operational){};
	\node[block, left of=below operational, node distance=5.2em](sotif){SOTIF\\(ISO~21448)};
	\node[block, left of=sotif, node distance=10.4em](functional){Functional safety\\(ISO~26262)};
	\node[block, right of=below operational, node distance=5.2em](behavioral){Behavioral safety};
	\node[block, fill=white, draw, right of=behavioral, node distance=10.4em](practice){Best practices\\(guidelines,\\ experience)};
	
	\node[coordinate, below of=operational, node distance=3.5em](operational helper){};
	\draw (operational) -- (operational helper) -| (functional);
	\draw (operational) -- (operational helper) -| (sotif);
	\draw (operational) -- (operational helper) -| (behavioral);
	\draw (operational) -- (operational helper) -| (practice);
\end{tikzpicture}}%

%% file: sections/assessment_pipeline.tex
\section{The safety assessment pipeline}
\label{sec:theSafetyAssessmentPipeline}

\begin{figure}[tb]
    \centering
    \input{figures/safety_assessment_pipeline.tikz}
    \caption{Outline of the safety assessment pipeline.}
    \label{fig:safetyAssessmentPipeline}
\end{figure}
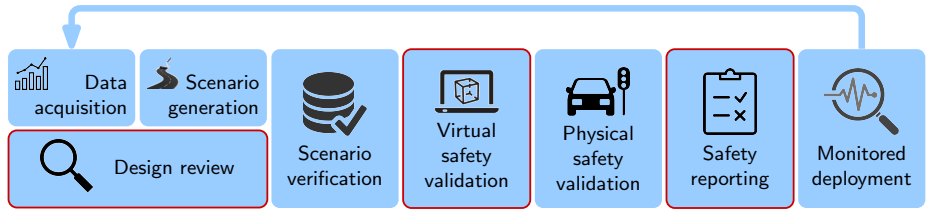

The proposed AV safety assessment methodology is schematically outlined in \cref{fig:safetyAssessmentPipeline}. This figure shows the assessment pipeline, which consists of eight steps (or `components') as further introduced below.

\begin{enumerate}
    \item\emph{Design review}\lsep
    During AV development, the AV developer must create a so-called \emph{safety case} according to ISO\,26262 standard on functional safety \cite{ISO26262}, which provides extensive information about the safety measures of the AV's automation system. This safety case serves a two-fold purpose in the scope of the assessment pipeline. First, the information contained in the safety case documents allows for a qualitative assessment of the functional safety of the AV, upon which a decision is made about the continuation of the safety assessment process. Second, the design review also yields insight into test scenarios that may be particularly challenging for the AV, or components of the automation system that pose a threat in case they fail (e.g., a `single point of failure'). This knowledge, in turn, is helpful in defining and selecting the test scenarios to be simulated in the \emph{scenario verification} component of the assessment pipeline.
    \item\emph{Data acquisition}\lsep
    This component aims to obtain real-world driving data, ultimately providing a statistically accurate set of traffic scenarios relevant to the AV. Ideally, the driving data should be obtained by AVs, since manual traffic may behave differently in the presence of AVs, but this is not feasible in the early stages of AV deployment. The data includes, among others, ego vehicle data and object-level data of other traffic participants. The data acquisition involves converting the collected data to a standardized format, allowing for a smooth transition to the next step of the pipeline, and also includes data validation and `cleaning' (e.g., removal of outliers).
    \item\emph{Scenario generation}\lsep
    In this step, candidate test scenarios are generated from the acquired data for virtual evaluation of the AV performance. To this end, measured scenarios are decomposed into elementary `events' and 'activities', enabling parametrization of the scenario data. Next, with a sufficient amount of measurement data, probability distributions of the scenario parameters are estimated. From these probability distributions, parameter sets can be sampled to obtain candidate traffic scenarios for virtual testing by re-creating variations of the original scenario. Here, safety-critical test scenarios are generated through parameter sampling which is biased towards safety-critical situations, i.e., the tails of the probability distribution.
    \item\emph{Scenario verification}\lsep
    Scenario verification involves reducing the number of candidate test scenarios by retaining only those scenarios that are categorized as being safety-critical for the particular AV under test. This component takes the results of both the \emph{scenario generation} and the \emph{design review} components as inputs, while also taking into account the so-called `safety of the intended functionality' (SOTIF) \cite{ISO21448}, and behavioral safety. On the other hand, the amount of test scenarios will be increased by introducing additional tests involving component failures, as dictated by functional safety considerations. The net result of this component is, therefore, a definitive set of test scenarios that provides quantitative information about all aspects of the operational safety of the AV when it is subjected to those test scenarios.  
    \item\emph{Virtual safety validation}\lsep
    The next component of the pipeline involves quantitative assessment of the AV safety through simulation of the selected test scenarios, employing a high-fidelity simulation environment capable of simulating the AV, its perception and control software, other road users, and the static environment. The simulation results are quantitatively assessed by means of key performance indicators (KPIs). These KPIs are subject to pass/fail thresholds, upon which it can be objectively decided whether to proceed to the next step of the assessment process or not. 
    \item\emph{Physical safety validation}\lsep
    This component of the assessment pipeline aims to verify the AV performance through practical tests, employing functional safety and SOTIF requirements, formulated in terms of KPIs. The first aspect of the physical safety validation involves the selection of a subset of the virtual test scenarios to perform physical testing, which includes tests with the AV on a test track as well as field tests in relevant real-world situations. Similar to the virtual safety validation component, quantitative assessment of the AV safety performance is performed through KPIs with pass/fail criteria, thus providing an objective basis for approval regarding deployment of the AV in the intended operational environment.
    \item\emph{Safety reporting}\lsep
    In this step, a comprehensive qualitative judgment, reflecting the overall impression, and a quantitative judgment, presenting the KPI values of previous assessment steps, are provided to the relevant authorities and the AV developer. Also, formal recommendations are given regarding the road approval of the AV to the relevant authorities.
    \item\emph{Monitored deployment}\lsep
    After successful completion of the previous component, the AV is allowed to drive on public roads, possibly restricted to certain areas and/or certain conditions. During this deployment phase, the AV developer is required to upload driving data in a near real-time fashion to allow for monitoring the AV behavior. This is implemented for two reasons. First, also after completion of the assessment pipeline, road authorities need to be able to monitor safety continuously. Second, the uploaded data is, after anonymization, fed back to the data acquisition component. The latter allows for continuous extension and improvement of the scenario database, while also being able to adapt to new types of transportation such as personal mobility devices.   
\end{enumerate}

\begin{remark}
    It is noted that the design review and the virtual safety validation include rating the AV safety level by means of pass/fail criteria, upon which the pipeline process will be continued or not. This is indicated in \cref{fig:safetyAssessmentPipeline} by means of solid red borders of the corresponding blocks. The physical safety validation, in fact, also includes a pass/fail rating, but the outcome will only be published in the safety report as the last step in the procedure.   
\end{remark}

%% file: figures/safety_assessment_pipeline.tikz
\resizebox{\textwidth}{!}{%
\begin{tikzpicture}
	\node[doublehalfblock, draw=TNOred, thick](dr) at (0,0) {%
        \begin{tabular}{@{}m{3em}@{\hspace{0.5em}}m{1.88\nwidth-3em-0.5em-2.1em}@{}}
            \includegraphics[width=2.5em]{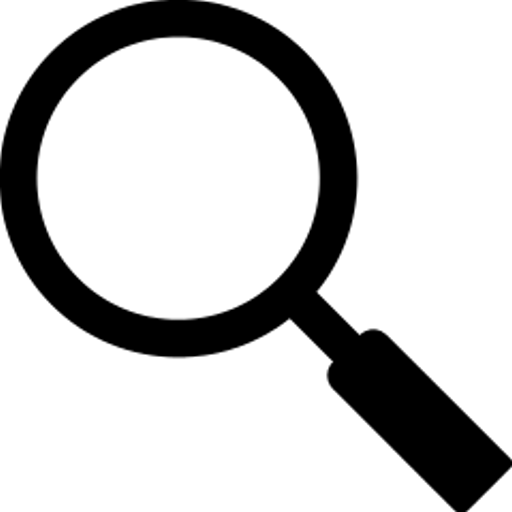} & Design review\\
        \end{tabular}};
    \node[halfblock, align=right](da) at (0,0.5\nheight+0.5\nsep) {%
    	\includegraphics[width=1.5em]{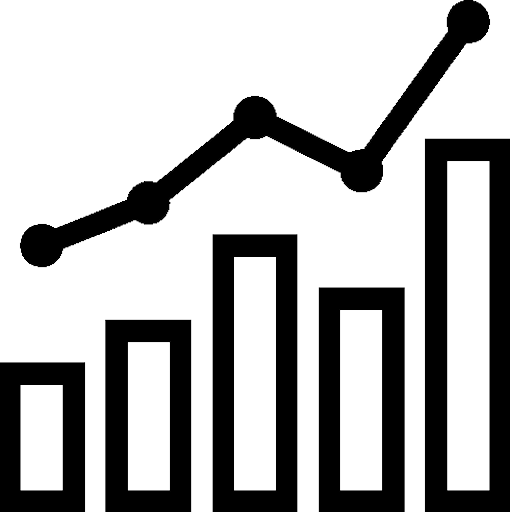}\hfill Data acquisition
    };
    \node[halfblock, align=right](tsg) at (\nwidth+\nsep,0.5\nheight+0.5\nsep) {%
    	\includegraphics[width=1.5em]{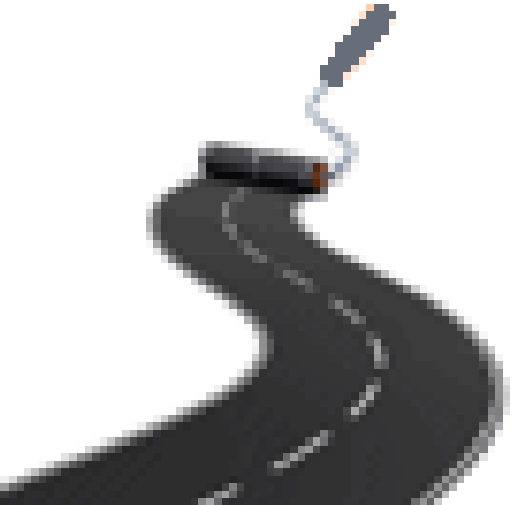}\hfill Scenario generation
    };
    \node[block](sv) at (2\nwidth+2\nsep,0) {%
    	\includegraphics[width=3em]{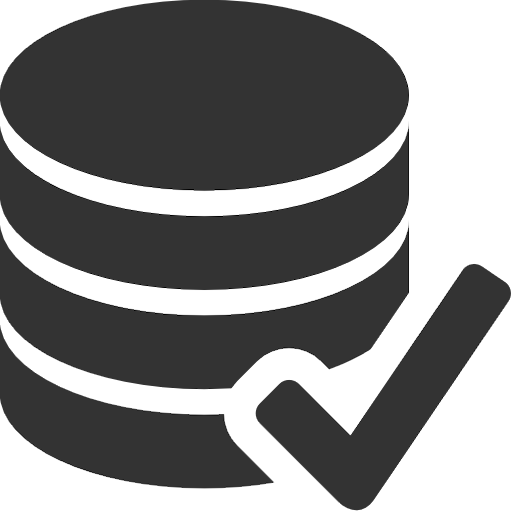}\\Scenario verification
    };
    \node[block, draw=TNOred, thick, right of=sv](vt){%
    	\includegraphics[width=3em]{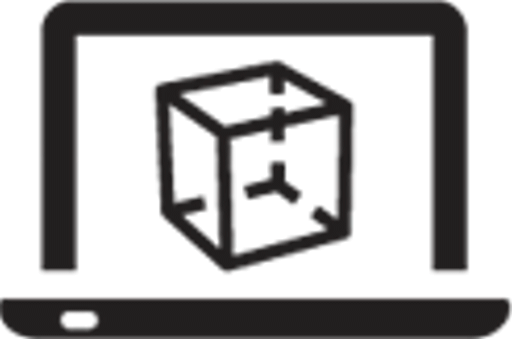}\\Virtual safety validation
    };
    \node[block, right of=vt](pt){%
    	\includegraphics[width=3em]{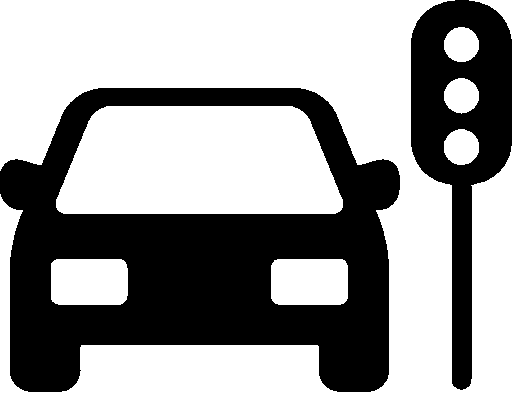}\\Physical safety validation
    };
    \node[block, draw=TNOred, thick, right of=pt](sr){%
    	\includegraphics[width=2.5em]{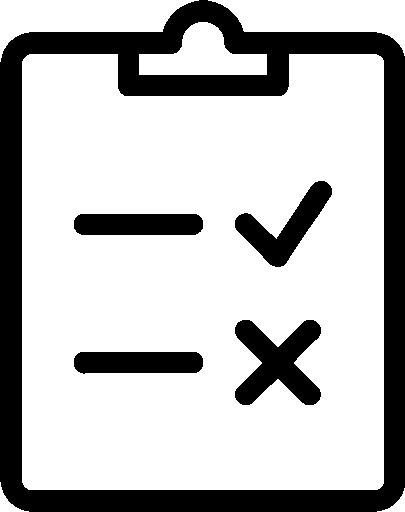}\\Safety reporting
    };
    \node[block, right of=sr](md){%
    	\includegraphics[width=3.5em]{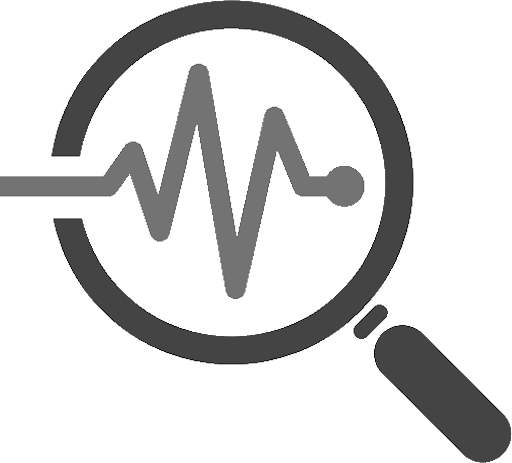}\\Monitored deployment
    };

	\node[coordinate, above of=da, node distance=@{\nheight/2}](helper feedback){};
	\draw[-{Triangle[width=3mm, length=5mm, round]}, line width=1mm, TNOlightblue, rounded corners=0.4em] (md) |- (helper feedback) -- (da);
\end{tikzpicture}}%

%% file: sections/scenario_generation.tex
\section{Scenario generation}
\label{sec:scenarioGeneration}

The objective of the scenario generation component is to generate candidate traffic scenarios for both virtual testing and physical testing with the ultimate goal to objectively and reproducibly assess operational safety of the AV\footnote{Note that the generated scenarios are still `candidate' scenarios. Only in the \emph{scenario verification} component, the definitive test scenarios are selected.}. The scenario generation is based on a unified ontology for road traffic, in which road-traffic \emph{scenarios} is an important notion, as further explained in the next two subsections.

\subsection{Scenarios}
\label{subsec:scenarios}

The scenario-based approach is the key property of the assessment methodology, which is why a definition of the notion of scenario is presented first. Several definitions of the notion of scenario in the context of automated driving are proposed \cite{Geyer2014, Ulbrich2015, Elrofai2016, Gelder2018}. We adopt the definition of De Gelder \etal\cite{Gelder2018}, as it is more applicable for the context of this paper. The definition of scenario is as follows.

\begin{definition}[Scenario] \label{def:scenario}
	A scenario is a quantitative description of the ego vehicle, its activities and/or goals, its static environment, and its dynamic environment. From the perspective of the ego vehicle, it contains all relevant events.
\end{definition}

Note that this definition is closely related to the definitions of Geyer \etal\cite{Geyer2014}, Ulbrich \etal\cite{Ulbrich2015} and Elrofai \etal\cite{Elrofai2016}. \Cref{def:scenario} deviates from these existing definitions in the fact that it explicitly states that a scenario is a quantitative description. Similar (quantitative) scenarios can be abstracted by means of a qualitative description, referred to as \emph{scenario class} \cite{Gelder2018}. A scenario class thus encompasses multiple scenarios.

\begin{figure}[tb]
    \centering
    \subfloat[Tags for weather condition, see \cite{Mahmassani2012}.]{%
        \small\input{figures/tree_weather.tikz}%
        \label{subfig:tree_weather}}\\
    \subfloat[Tags for road type, inspired by \cite{Bonnin2014}.]{%
        \small\input{figures/tree_road.tikz}%
        \label{subfig:tree_road}}\\
    \subfloat[Tags for target maneuvers, see \cite{Gelder2018}.]{%
        \small\input{figures/tree_target.tikz}%
        \label{subfig:tree_target}}%
    \caption{Examples of hierarchical structures (`trees') of tags regarding \protect\subref{subfig:tree_weather} type of weather, \protect\subref{subfig:tree_road} road type, and \protect\subref{subfig:tree_target} target maneuvers.}
    \label{fig:tagTrees}
\end{figure}
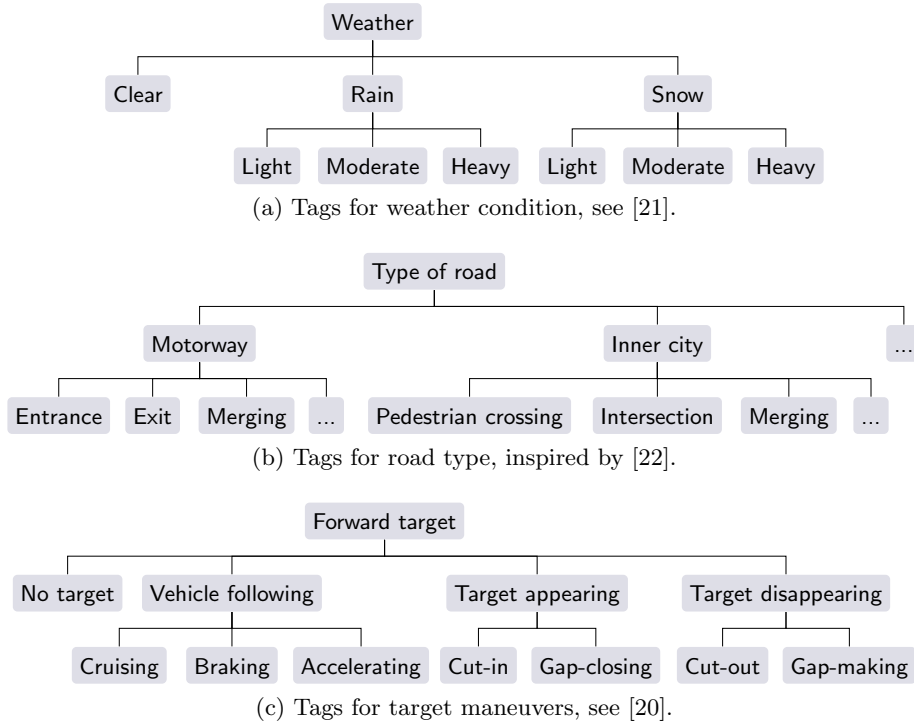

To store measured scenarios in a structured way, such that they can be easily retrieved afterwards for test scenario generation, it is proposed to set up a hierarchical system of \emph{tags}, corresponding to the aforementioned road traffic ontology\footnote{Alternatively, a hierarchical system of types of scenarios may be created, thus directly classifying the scenario itself. However, scenarios may not be mutually exclusive. For instance, the same scenario might be relevant in highway and in urban environments. In case of direct scenario classification, this situation can only be solved by storing the same scenario multiple times, which is inefficient and prone to errors.}. As such, tags provide a semantic classification of the scenario in terms of, e.g., weather, road type, and target vehicle maneuvers. In accordance with the hierarchical approach, tags are arranged into \emph{trees}, as illustrated in \cref{fig:tagTrees}. This figure shows three examples, relating to type of weather, road type, and target vehicle maneuver, respectively. Tags that are in the same layer are mutually exclusive, whereas different layers of a tree represent different levels of abstraction \cite{Bonnin2014}.

\subsection{Events}
\label{subsec:events}

A scenario can be decomposed into elementary building blocks, so-called \emph{events} and \emph{activities}. An event is defined as follows \cite{Gelder2018}.

\begin{definition}[Event]\label{def:event}
	An event marks the time instant at which a mode transition occurs, such that before and after an event, the state corresponds to two different modes.
\end{definition}

A mode transition may be caused by an abrupt change of either an input signal, a parameter, or a state. For example, pushing the brake pedal may cause a mode transition and therefore, this may be regarded as an event. The inter-event time interval, i.e., the time period between two subsequent events, is characterized by the occurrence of a certain \emph{activity}. For example, when the longitudinal acceleration is negative during such an inter-event time interval, the activity corresponds to braking. 

\begin{figure}[tb]
	\centering
	\setlength\figurewidth{1.11\textwidth}
	\setlength\figureheight{180pt}
	\input{figures/events.tikz}
	\caption{Example of a scenario decomposition into parallel and serial activities. The vertical lines represent the events. The red vertical line at $t=16\,\text{s}$ indicates the end of the first scenario and the start of the second scenario.}
	\label{fig:eventsExample}
\end{figure}
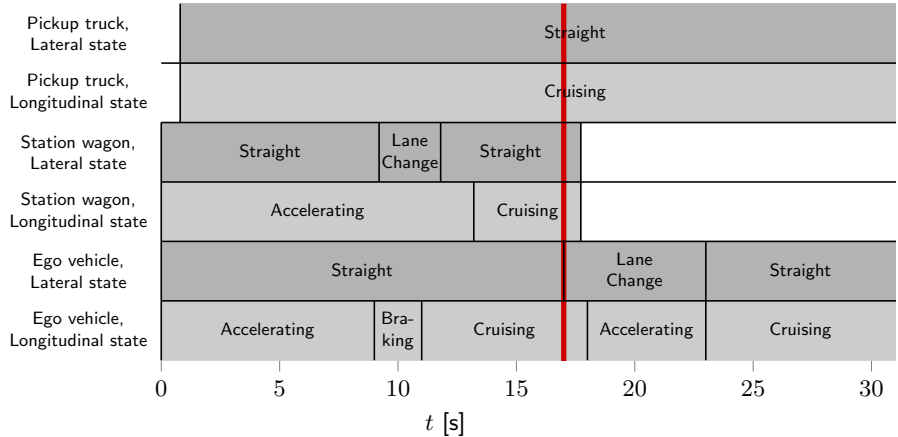

In general, a scenario contains multiple events and activities, whose temporal properties are such that both parallel and serial activities can be distinguished. As an example, \cref{fig:exampleSchematic} shows two consecutive scenarios in which the ego vehicle (the red vehicle) is being overtaken by a station wagon (first scenario), after which the ego vehicle overtakes a pickup truck (second scenario). \Cref{fig:eventsExample} shows the activities and events for this example, in which the behaviors of the three actors are characterized by a number of serial activities for the longitudinal and lateral behavior. Events and activities thus relate to actors.

\begin{figure}[tb]
	\centering
	\setlength\figureheight{106pt}
	\setlength\figurewidth{260pt}
	\subfloat[Last vehicle overtakes other vehicles.]{\input{figures/scenario_situation_1.tikz}}\\
	\setlength\figureheight{100pt}
	\subfloat[Another overtaking.]{\input{figures/scenario_situation_2.tikz}}
	\caption{Schematic overview of two consecutive traffic scenarios. The sedan (red vehicle) is defined as the ego vehicle. First, the station wagon (the last vehicle in (a)) overtakes both vehicles. Next, the ego vehicle overtakes the pickup truck (first vehicle in (a)).}
	\label{fig:exampleSchematic}
\end{figure}
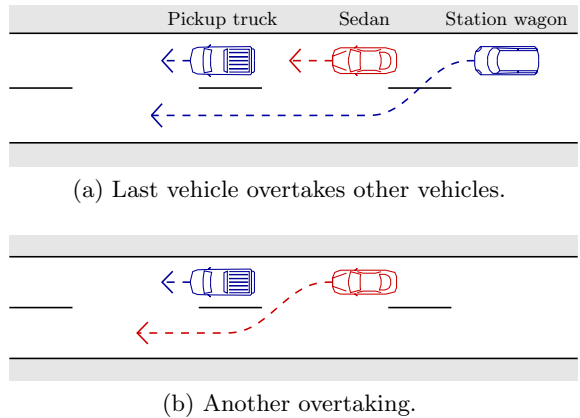

The advantage of decomposing scenarios into elementary events and (inter-event) activities is threefold:
\begin{enumerate}
	\item Whereas the amount of relevant scenarios might become very large over time, it is expected that only a \emph{limited set of essentially different activities} will be obtained, being a direct consequence of the rather limited number of actions a vehicle can perform (basically change of speed and direction).
	\item When analyzing measured data to extract scenarios, events are instrumental in \emph{automatic scenario recognition}. Similarly, events can be directly detected by the ego vehicle in real life, providing a trigger for data logging.
	\item The decomposition of a scenario into several activities allows for parametrization of scenarios, which, in turn, enables estimation of a (possibly multivariate) probability density function (PDF) of the parameters. This PDF can then be sampled in a specific way, e.g., emphasizing safety-critical situations or generating activities that are not actually encountered in the measurement data. Moreover, sampling from the PDF creates the possibility to automatically sample those situations from the parameter set during which the vehicle under test shows safety-critical behavior, without \textit{a priori} knowledge of which scenarios are critical for the vehicle \cite{Gelder2017}. In other words, scenario parametrization allows for the systematic generation of candidate test scenarios for safety assessment of the vehicle under test.
\end{enumerate}

\subsection{Scenario generation topics}
\label{subsec:scenarioGenerationTopics}

Having introduced the notions of scenario, event, and activity, the scenario generation component of the assessment pipeline can now be described in terms of the different topics below, adding more detail to the brief description in \cref{sec:theSafetyAssessmentPipeline}.
\begin{enumerate}
	\item\emph{Event detection}\lsep This involves the automated detection of events, either during off-line analysis of measurement data, or in real-time during data collection. Since the inter-event time intervals relate to the activities, the detection of events automatically result in the detection of the inter-event activities. 
	\item\emph{Activity parametrization}\lsep Next, detected activities will be parametrized and the PDF of the parameters will be estimated.
	\item\emph{Scenario mining}\lsep Next, based on certain `patterns' in detected activities and events, they will be clustered into full scenarios. This topic also includes scenario tagging.
	\item\emph{Test scenario generation}\lsep Finally, (candidate) test scenarios can be generated by selecting scenario classes of interest, using the tagging system, upon which instances of those scenario classes are generated by drawing samples from the probability distribution of the activity parameters\footnote{It is still a topic of research which and how many test scenarios need to be sampled from the total set of real-world scenarios. These questions are covered by the notion of \emph{completeness}.}. Here, an `instance of a scenario class' is the quantitative scenario description. As explained in \cref{subsec:events}, the sampling is biased towards the tails of the probability distribution, representing the most extreme and/or rare situations.
\end{enumerate}

%% file: figures/tree_weather.tikz
\definecolor{TNOlightgray}{RGB}{222,222,231}%
\tikzstyle{tag}=[text height=.8em, text depth=.1em, font=\sffamily, rounded corners=0.2em, fill=TNOlightgray]%
\tikzstyle{helper}=[coordinate, node distance=1.5em]%
\scalebox{0.96}{%
\begin{tikzpicture}
	\node[tag](weather){Weather};
	\node[tag, below of=weather](rain){Rain};
	\node[tag, left of=rain, node distance=10em](clear){Clear};
	\node[tag, right of=rain, node distance=13em](snow){Snow};
	\node[tag, below of=rain](mod rain){Moderate};
	\node[tag, left of=mod rain, node distance=4.5em](light rain){Light};
	\node[tag, right of=mod rain, node distance=4.6em](heavy rain){Heavy};
	\node[tag, below of=snow](mod snow){Moderate};
	\node[tag, left of=mod snow, node distance=4.5em](light snow){Light};
	\node[tag, right of=mod snow, node distance=4.6em](heavy snow){Heavy};
	
	\node[helper, below of=weather](weather helper){};
	\node[helper, below of=rain](rain helper){};
	\node[helper, below of=snow](snow helper){};
	\draw (weather) -- (rain);
	\draw (weather) -- (weather helper) -| (clear);
	\draw (weather) -- (weather helper) -| (snow);
	\draw (rain) -- (mod rain);
	\draw (rain) -- (rain helper) -| (light rain);
	\draw (rain) -- (rain helper) -| (heavy rain);
	\draw (snow) -- (mod snow);
	\draw (snow) -- (snow helper) -| (light snow);
	\draw (snow) -- (snow helper) -| (heavy snow);
\end{tikzpicture}}%

%% file: figures/tree_road.tikz
\definecolor{TNOlightgray}{RGB}{222,222,231}%
\tikzstyle{tag}=[text height=.8em, text depth=.1em, font=\sffamily, rounded corners=0.2em, fill=TNOlightgray]%
\tikzstyle{helper}=[coordinate, node distance=1.5em]%
\scalebox{0.96}{%
\begin{tikzpicture}
	\node[tag](road){Type of road};
	\node[coordinate, below of=road](below road){};
	\node[tag, left of=below road, node distance=10em](motorway){Motorway};
	\node[tag, right of=below road, node distance=9.5em](inner city){Inner city};
	\node[tag, right of=inner city, node distance=10.5 em](remaining a){...};
	\node[coordinate, below of=motorway](below motorway){};
	\node[tag, left of=below motorway, node distance=2em](exit){Exit};
	\node[tag, left of=exit, node distance=4em](entrance){Entrance};
	\node[tag, right of=below motorway, node distance=2em](merging motorway){Merging};
	\node[tag, right of=merging motorway, node distance=3.4em](remaining b){...};
	\node[tag, below of=inner city](intersection){Intersection};
	\node[tag, left of=intersection, node distance=8em](pedestrian crossing){Pedestrian crossing};
	\node[tag, right of=intersection, node distance=5.6em](merging inner city){Merging};
	\node[tag, right of=merging inner city, node distance=3.5em](remaining c){...};
	
	\node[helper, below of=road](road helper){};
	\node[helper, below of=motorway](motorway helper){};
	\node[helper, below of=inner city](inner city helper){};
	\draw (road) -- (road helper) -| (motorway);
	\draw (road) -- (road helper) -| (inner city);
	\draw (road) -- (road helper) -| (remaining a);
	\draw (motorway) -- (motorway helper) -| (entrance);
	\draw (motorway) -- (motorway helper) -| (exit);
	\draw (motorway) -- (motorway helper) -| (merging motorway);
	\draw (motorway) -- (motorway helper) -| (remaining b);
	\draw (inner city) -- (inner city helper) -| (pedestrian crossing);
	\draw (inner city) -- (intersection);
	\draw (inner city) -- (inner city helper) -| (merging inner city);
	\draw (inner city) -- (inner city helper) -| (remaining c);
\end{tikzpicture}}%

%% file: figures/tree_target.tikz
\definecolor{TNOlightgray}{RGB}{222,222,231}%
\tikzstyle{tag}=[text height=.8em, text depth=.1em, font=\sffamily, rounded corners=0.2em, fill=TNOlightgray]%
\tikzstyle{helper}=[coordinate, node distance=1.5em]%
\scalebox{0.96}{%
\begin{tikzpicture}
	\node[tag](target){Forward target};
	\node[coordinate, below of=target](below target){};
	\node[tag, left of=below target, node distance=6.5em](following){Vehicle following};
	\node[tag, left of=following, node distance=7em](free){No target};
	\node[tag, right of=below target, node distance=6.5em](appearing){Target appearing};
	\node[tag, right of=appearing, node distance=10.6em](disappearing){Target disappearing};
	\node[tag, below of=following](braking){Braking};
	\node[tag, left of=braking, node distance=4.8em](cruising){Cruising};
	\node[tag, right of=braking, node distance=5.5em](accelerating){Accelerating};
	\node[coordinate, below of=appearing](below appearing){};
	\node[tag, left of=below appearing, node distance=2.5em](cutin){Cut-in};
	\node[tag, right of=below appearing, node distance=2.5em](gapclosing){Gap-closing};
	\node[coordinate, below of=disappearing](below disappearing){};
	\node[tag, left of=below disappearing, node distance=2.7em](cutout){Cut-out};
	\node[tag, right of=below disappearing, node distance=2.7em](gapmaking){Gap-making};
	
	\node[helper, below of=target](target helper){};
	\node[helper, below of=following](following helper){};
	\node[helper, below of=appearing](appearing helper){};
	\node[helper, below of=disappearing](disappearing helper){};
	\draw (target) -- (target helper) -| (free);
	\draw (target) -- (target helper) -| (following);
	\draw (target) -- (target helper) -| (appearing);
	\draw (target) -- (target helper) -| (disappearing);
	\draw (following) -- (following helper) -| (cruising);
	\draw (following) -- (braking);
	\draw (following) -- (following helper) -| (accelerating);
	\draw (appearing) -- (appearing helper) -| (cutin);
	\draw (appearing) -- (appearing helper) -| (gapclosing);
	\draw (disappearing) -- (disappearing helper) -| (cutout);
	\draw (disappearing) -- (disappearing helper) -| (gapmaking);
\end{tikzpicture}}%

%% file: figures/scenario_situation_1.tikz

\definecolor{TNOred}{RGB}{204,0,0}
\definecolor{TNOblue}{RGB}{0,0,153}

\begin{tikzpicture}

\begin{axis}[
xmin=-20, xmax=20,
ymin=-5, ymax=5.3,
width=\figurewidth,
height=\figureheight,
tick align=outside,
tick pos=left,
x grid style={white!69.019607843137251!black},
y grid style={white!69.019607843137251!black},
axis background/.style={fill=white!90.0!black},
ticks=none,
hide axis
]
\path [draw=white, fill=white] (axis cs:-20,3.5)
--(axis cs:20,3.5)
--(axis cs:20,-3.5)
--(axis cs:-20,-3.5)
--cycle;

\addplot [semithick, black, forget plot]
table {%
-20 3.5
20 3.5
};
\addplot [semithick, black, forget plot]
table {%
-20 -3.5
20 -3.5
};
\addplot [semithick, black, forget plot]
table {%
-20 0
-15.5555555555556 0
};
\addplot [semithick, black, forget plot]
table {%
-6.66666666666666 0
-2.22222222222222 0
};
\addplot [semithick, black, forget plot]
table {%
6.66666666666667 0
11.1111111111111 0
};
\addplot [TNOred, forget plot]
table {%
7.25 1.77571428571429
7.25 1.36428571428571
7.13235294117647 1.08142857142857
6.48529411764706 0.927142857142857
4.33823529411765 0.927142857142857
4.48529411764706 0.798571428571428
4.33823529411765 0.927142857142857
3.30882352941176 0.927142857142857
2.89705882352941 1.08142857142857
2.75 1.44142857142857
2.75 1.75
2.75 1.46714285714286
3.72058823529412 1.13285714285714
};
\addplot [TNOred, forget plot]
table {%
7.25 1.72428571428571
7.25 2.13571428571429
7.13235294117647 2.41857142857143
6.48529411764706 2.57285714285714
4.33823529411765 2.57285714285714
4.48529411764706 2.70142857142857
4.33823529411765 2.57285714285714
3.30882352941176 2.57285714285714
2.89705882352941 2.41857142857143
2.75 2.05857142857143
2.75 1.75
2.75 2.03285714285714
3.72058823529412 2.36714285714286
};
\addplot [TNOred, forget plot]
table {%
6.16176470588235 1.03
5.77941176470588 1.13285714285714
4.89705882352941 1.13285714285714
4.48529411764706 1.03
};
\addplot [TNOred, forget plot]
table {%
6.16176470588235 2.47
5.77941176470588 2.36714285714286
4.89705882352941 2.36714285714286
4.48529411764706 2.47
};
\addplot [TNOred, forget plot]
table {%
6.16176470588235 1.75
6.16176470588235 1.57
6.04411764705882 1.21
6.39705882352941 1.21
6.63235294117647 1.28714285714286
6.75 1.39
6.83823529411765 1.59571428571429
6.83823529411765 1.75
};
\addplot [TNOred, forget plot]
table {%
6.16176470588235 1.75
6.16176470588235 1.93
6.04411764705882 2.29
6.39705882352941 2.29
6.63235294117647 2.21285714285714
6.75 2.11
6.83823529411765 1.90428571428571
6.83823529411765 1.75
};
\addplot [TNOred, forget plot]
table {%
4.69117647058824 1.75
4.69117647058824 1.41571428571429
4.72058823529412 1.21
4.01470588235294 1.03
3.86764705882353 1.62142857142857
3.86764705882353 1.75
};
\addplot [TNOred, forget plot]
table {%
4.69117647058824 1.75
4.69117647058824 2.08428571428571
4.72058823529412 2.29
4.01470588235294 2.47
3.86764705882353 1.87857142857143
3.86764705882353 1.75
};
\addplot [TNOblue, forget plot]
table {%
-3.22093023255814 1.75
-3.22093023255814 1.73767123287671
-4.50290697674419 1.73767123287671
-4.50290697674419 1.75
};
\addplot [TNOblue, forget plot]
table {%
-3.22093023255814 1.75
-3.22093023255814 1.76232876712329
-4.50290697674419 1.76232876712329
-4.50290697674419 1.75
};
\addplot [TNOblue, forget plot]
table {%
-3.22093023255814 1.49109589041096
-4.50290697674419 1.49109589041096
-4.50290697674419 1.51575342465753
-3.22093023255814 1.51575342465753
-3.22093023255814 1.49109589041096
};
\addplot [TNOblue, forget plot]
table {%
-3.22093023255814 2.00890410958904
-4.50290697674419 2.00890410958904
-4.50290697674419 1.98424657534247
-3.22093023255814 1.98424657534247
-3.22093023255814 2.00890410958904
};
\addplot [TNOblue, forget plot]
table {%
-3.22093023255814 1.21986301369863
-4.50290697674419 1.21986301369863
-4.50290697674419 1.24452054794521
-3.22093023255814 1.24452054794521
-3.22093023255814 1.21986301369863
};
\addplot [TNOblue, forget plot]
table {%
-3.22093023255814 2.28013698630137
-4.50290697674419 2.28013698630137
-4.50290697674419 2.25547945205479
-3.22093023255814 2.25547945205479
-3.22093023255814 2.28013698630137
};
\addplot [TNOblue, forget plot]
table {%
-2.75 1.75
-2.75 1.07191780821918
-2.82848837209302 0.997945205479452
-2.9593023255814 0.997945205479452
-2.9593023255814 1.75
-2.9593023255814 0.923972602739726
-3.03779069767442 0.85
-5.99418604651163 0.874657534246575
-5.83720930232558 0.726712328767123
-5.99418604651163 0.874657534246575
-6.75290697674419 0.899315068493151
-7.09302325581395 1.02260273972603
-7.25 1.63904109589041
-7.25 1.75
};
\addplot [TNOblue, forget plot]
table {%
-2.75 1.75
-2.75 2.42808219178082
-2.82848837209302 2.50205479452055
-2.9593023255814 2.50205479452055
-2.9593023255814 1.75
-2.9593023255814 2.57602739726027
-3.03779069767442 2.65
-5.99418604651163 2.62534246575343
-5.83720930232558 2.77328767123288
-5.99418604651163 2.62534246575343
-6.75290697674419 2.60068493150685
-7.09302325581395 2.47739726027397
-7.25 1.86095890410959
-7.25 1.75
};
\addplot [TNOblue, forget plot]
table {%
-4.65988372093023 1.75
-4.65988372093023 0.997945205479452
-4.8953488372093 1.09657534246575
-4.8953488372093 1.75
};
\addplot [TNOblue, forget plot]
table {%
-4.65988372093023 1.75
-4.65988372093023 2.50205479452055
-4.8953488372093 2.40342465753425
-4.8953488372093 1.75
};
\addplot [TNOblue, forget plot]
table {%
-5.83720930232558 1.75
-5.83720930232558 1.44178082191781
-5.75872093023256 1.07191780821918
-6.28197674418605 0.973287671232877
-6.36046511627907 1.46643835616438
-6.36046511627907 1.75
};
\addplot [TNOblue, forget plot]
table {%
-5.83720930232558 1.75
-5.83720930232558 2.05821917808219
-5.75872093023256 2.42808219178082
-6.28197674418605 2.52671232876712
-6.36046511627907 2.03356164383562
-6.36046511627907 1.75
};
\addplot [TNOblue, forget plot]
table {%
-3.11627906976744 1.75
-3.11627906976744 1.02260273972603
-4.52906976744186 1.02260273972603
};
\addplot [TNOblue, forget plot]
table {%
-3.11627906976744 1.75
-3.11627906976744 2.47739726027397
-4.52906976744186 2.47739726027397
};
\addplot [TNOblue, forget plot]
table {%
17.25 1.75
17.25 1.40714285714286
17.1590909090909 1.10714285714286
14.5227272727273 1.10714285714286
14.3863636363636 1.40714285714286
14.3409090909091 1.53571428571429
14.3409090909091 1.75
};
\addplot [TNOblue, forget plot]
table {%
17.25 1.75
17.25 2.09285714285714
17.1590909090909 2.39285714285714
14.5227272727273 2.39285714285714
14.3863636363636 2.09285714285714
14.3409090909091 1.96428571428571
14.3409090909091 1.75
};
\addplot [TNOblue, forget plot]
table {%
14.5227272727273 1.10714285714286
14.2954545454545 0.935714285714286
14.0227272727273 0.935714285714286
13.7954545454545 1.15
13.6590909090909 1.49285714285714
13.6590909090909 1.75
};
\addplot [TNOblue, forget plot]
table {%
14.5227272727273 2.39285714285714
14.2954545454545 2.56428571428571
14.0227272727273 2.56428571428571
13.7954545454545 2.35
13.6590909090909 2.00714285714286
13.6590909090909 1.75
};
\addplot [TNOblue, forget plot]
table {%
14.2954545454545 0.935714285714286
15.7954545454545 0.978571428571429
15.7954545454545 1.10714285714286
15.7954545454545 0.978571428571429
17.1136363636364 0.978571428571429
17.1590909090909 1.10714285714286
17.1136363636364 0.978571428571429
16.9772727272727 0.85
13.1590909090909 0.85
12.8409090909091 1.15
13.0227272727273 1.15
13.2045454545455 0.978571428571429
13.2045454545455 0.85
};
\addplot [TNOblue, forget plot]
table {%
14.2954545454545 2.56428571428571
15.7954545454545 2.52142857142857
15.7954545454545 2.39285714285714
15.7954545454545 2.52142857142857
17.1136363636364 2.52142857142857
17.1590909090909 2.39285714285714
17.1136363636364 2.52142857142857
16.9772727272727 2.65
13.1590909090909 2.65
12.8409090909091 2.35
13.0227272727273 2.35
13.2045454545455 2.52142857142857
13.2045454545455 2.65
};
\addplot [TNOblue, forget plot]
table {%
12.8409090909091 1.15
12.75 1.45
12.75 1.75
};
\addplot [TNOblue, forget plot]
table {%
12.8409090909091 2.35
12.75 2.05
12.75 1.75
};
\addplot [semithick, TNOblue, dashed, forget plot]
table {%
12.75 1.75
12.6717171717172 1.74996447297214
12.5934343434343 1.74972010306519
12.5151515151515 1.74906980619819
12.4368686868687 1.74782908517264
12.3585858585859 1.74582576468553
12.280303030303 1.74289972634234
12.2020202020202 1.73890264367003
12.1237373737374 1.73369771713003
12.0454545454545 1.72715940913127
11.9671717171717 1.71917317904316
11.8888888888889 1.70963521820861
11.8106060606061 1.69845218495698
11.7323232323232 1.68554093961714
11.6540404040404 1.67082827953046
11.5757575757576 1.65425067406376
11.4974747474747 1.63575399962237
11.4191919191919 1.6152932746631
11.3409090909091 1.59283239470727
11.2626262626263 1.56834386735364
11.1843434343434 1.54180854729151
11.1060606060606 1.51321537131363
11.0277777777778 1.48256109332927
10.9494949494949 1.44985001937716
10.8712121212121 1.41509374263854
10.7929292929293 1.37831087845013
10.7146464646465 1.33952679931715
10.6363636363636 1.2987733699263
10.5580808080808 1.25608868215878
10.479797979798 1.21151679010327
10.4015151515152 1.16510744506897
10.3232323232323 1.11691583059853
10.2449494949495 1.06700229748114
10.1666666666667 1.01543209876543
10.0883838383838 0.962275124772575
10.010101010101 0.907605638109211
9.93181818181818 0.85150200868048
9.85353535353535 0.794046448703017
9.77525252525253 0.735324747717952
9.6969696969697 0.67542600760391
9.61868686868687 0.614442377590011
9.54040404040404 0.552468789268875
9.46212121212121 0.489602691609613
9.38383838383838 0.425943785970835
9.30555555555556 0.361593761113651
9.22727272727273 0.296656028214665
9.1489898989899 0.231235455878979
9.07070707070707 0.165438105153197
8.99242424242424 0.0993709645384162
8.91414141414141 0.033141685003238
8.83585858585858 -0.0331416850032387
8.75757575757576 -0.0993709645384178
8.67929292929293 -0.165438105153197
8.6010101010101 -0.23123545587898
8.52272727272727 -0.296656028214666
8.44444444444444 -0.361593761113652
8.36616161616162 -0.425943785970836
8.28787878787879 -0.489602691609613
8.20959595959596 -0.552468789268874
8.13131313131313 -0.614442377590012
8.0530303030303 -0.67542600760391
7.97474747474747 -0.735324747717952
7.89646464646465 -0.794046448703018
7.81818181818182 -0.851502008680481
7.73989898989899 -0.907605638109211
7.66161616161616 -0.962275124772574
7.58333333333333 -1.01543209876543
7.50505050505051 -1.06700229748113
7.42676767676768 -1.11691583059853
7.34848484848485 -1.16510744506897
7.27020202020202 -1.21151679010327
7.19191919191919 -1.25608868215878
7.11363636363636 -1.2987733699263
7.03535353535353 -1.33952679931715
6.95707070707071 -1.37831087845013
6.87878787878788 -1.41509374263854
6.80050505050505 -1.44985001937716
6.72222222222222 -1.48256109332927
6.64393939393939 -1.51321537131363
6.56565656565657 -1.54180854729151
6.48737373737374 -1.56834386735365
6.40909090909091 -1.59283239470727
6.33080808080808 -1.6152932746631
6.25252525252525 -1.63575399962237
6.17424242424242 -1.65425067406375
6.0959595959596 -1.67082827953045
6.01767676767677 -1.68554093961715
5.93939393939394 -1.69845218495698
5.86111111111111 -1.7096352182086
5.78282828282828 -1.71917317904316
5.70454545454545 -1.72715940913127
5.62626262626263 -1.73369771713003
5.5479797979798 -1.73890264367003
5.46969696969697 -1.74289972634234
5.39141414141414 -1.74582576468553
5.31313131313131 -1.74782908517264
5.23484848484848 -1.7490698061982
5.15656565656566 -1.74972010306519
5.07828282828283 -1.74996447297215
5 -1.75
-10 -1.75
};
\addplot [semithick, TNOblue, forget plot]
table {%
-10 -1.75
-9.25 -1
};
\addplot [semithick, TNOblue, forget plot]
table {%
-10 -1.75
-9.25 -2.5
};
\addplot [semithick, TNOred, dashed, forget plot]
table {%
2.75 1.75
-0.25 1.75
};
\addplot [semithick, TNOred, forget plot]
table {%
0.5 2.5
-0.25 1.75
0.5 1
};
\addplot [semithick, TNOblue, dashed, forget plot]
table {%
-7.25 1.75
-9.25 1.75
};
\addplot [semithick, TNOblue, forget plot]
table {%
-8.5 2.5
-9.25 1.75
-8.5 1
};
\node at (axis cs:5,5.3)[
  scale=0.8,
  anchor=north,
  text=black,
  rotate=0.0
]{ Sedan};
\node at (axis cs:-5,5.3)[
  scale=0.8,
  anchor=north,
  text=black,
  rotate=0.0
]{ Pickup truck};
\node at (axis cs:15,5.3)[
  scale=0.8,
  anchor=north,
  text=black,
  rotate=0.0
]{ Station wagon};
\end{axis}

\end{tikzpicture}

%% file: figures/scenario_situation_2.tikz

\definecolor{TNOred}{RGB}{204,0,0}
\definecolor{TNOblue}{RGB}{0,0,153}

\begin{tikzpicture}

\begin{axis}[
xmin=-20, xmax=20,
ymin=-5, ymax=5,
width=\figurewidth,
height=\figureheight,
tick align=outside,
tick pos=left,
x grid style={white!69.019607843137251!black},
y grid style={white!69.019607843137251!black},
axis background/.style={fill=white!90.0!black},
ticks=none,
hide axis
]
\path [draw=white, fill=white] (axis cs:-20,3.5)
--(axis cs:20,3.5)
--(axis cs:20,-3.5)
--(axis cs:-20,-3.5)
--cycle;

\addplot [semithick, black, forget plot]
table {%
-20 3.5
20 3.5
};
\addplot [semithick, black, forget plot]
table {%
-20 -3.5
20 -3.5
};
\addplot [semithick, black, forget plot]
table {%
-20 0
-15.5555555555556 0
};
\addplot [semithick, black, forget plot]
table {%
-6.66666666666666 0
-2.22222222222222 0
};
\addplot [semithick, black, forget plot]
table {%
6.66666666666667 0
11.1111111111111 0
};
\addplot [TNOred, forget plot]
table {%
7.25 1.77571428571429
7.25 1.36428571428571
7.13235294117647 1.08142857142857
6.48529411764706 0.927142857142857
4.33823529411765 0.927142857142857
4.48529411764706 0.798571428571428
4.33823529411765 0.927142857142857
3.30882352941176 0.927142857142857
2.89705882352941 1.08142857142857
2.75 1.44142857142857
2.75 1.75
2.75 1.46714285714286
3.72058823529412 1.13285714285714
};
\addplot [TNOred, forget plot]
table {%
7.25 1.72428571428571
7.25 2.13571428571429
7.13235294117647 2.41857142857143
6.48529411764706 2.57285714285714
4.33823529411765 2.57285714285714
4.48529411764706 2.70142857142857
4.33823529411765 2.57285714285714
3.30882352941176 2.57285714285714
2.89705882352941 2.41857142857143
2.75 2.05857142857143
2.75 1.75
2.75 2.03285714285714
3.72058823529412 2.36714285714286
};
\addplot [TNOred, forget plot]
table {%
6.16176470588235 1.03
5.77941176470588 1.13285714285714
4.89705882352941 1.13285714285714
4.48529411764706 1.03
};
\addplot [TNOred, forget plot]
table {%
6.16176470588235 2.47
5.77941176470588 2.36714285714286
4.89705882352941 2.36714285714286
4.48529411764706 2.47
};
\addplot [TNOred, forget plot]
table {%
6.16176470588235 1.75
6.16176470588235 1.57
6.04411764705882 1.21
6.39705882352941 1.21
6.63235294117647 1.28714285714286
6.75 1.39
6.83823529411765 1.59571428571429
6.83823529411765 1.75
};
\addplot [TNOred, forget plot]
table {%
6.16176470588235 1.75
6.16176470588235 1.93
6.04411764705882 2.29
6.39705882352941 2.29
6.63235294117647 2.21285714285714
6.75 2.11
6.83823529411765 1.90428571428571
6.83823529411765 1.75
};
\addplot [TNOred, forget plot]
table {%
4.69117647058824 1.75
4.69117647058824 1.41571428571429
4.72058823529412 1.21
4.01470588235294 1.03
3.86764705882353 1.62142857142857
3.86764705882353 1.75
};
\addplot [TNOred, forget plot]
table {%
4.69117647058824 1.75
4.69117647058824 2.08428571428571
4.72058823529412 2.29
4.01470588235294 2.47
3.86764705882353 1.87857142857143
3.86764705882353 1.75
};
\addplot [TNOblue, forget plot]
table {%
-3.22093023255814 1.75
-3.22093023255814 1.73767123287671
-4.50290697674419 1.73767123287671
-4.50290697674419 1.75
};
\addplot [TNOblue, forget plot]
table {%
-3.22093023255814 1.75
-3.22093023255814 1.76232876712329
-4.50290697674419 1.76232876712329
-4.50290697674419 1.75
};
\addplot [TNOblue, forget plot]
table {%
-3.22093023255814 1.49109589041096
-4.50290697674419 1.49109589041096
-4.50290697674419 1.51575342465753
-3.22093023255814 1.51575342465753
-3.22093023255814 1.49109589041096
};
\addplot [TNOblue, forget plot]
table {%
-3.22093023255814 2.00890410958904
-4.50290697674419 2.00890410958904
-4.50290697674419 1.98424657534247
-3.22093023255814 1.98424657534247
-3.22093023255814 2.00890410958904
};
\addplot [TNOblue, forget plot]
table {%
-3.22093023255814 1.21986301369863
-4.50290697674419 1.21986301369863
-4.50290697674419 1.24452054794521
-3.22093023255814 1.24452054794521
-3.22093023255814 1.21986301369863
};
\addplot [TNOblue, forget plot]
table {%
-3.22093023255814 2.28013698630137
-4.50290697674419 2.28013698630137
-4.50290697674419 2.25547945205479
-3.22093023255814 2.25547945205479
-3.22093023255814 2.28013698630137
};
\addplot [TNOblue, forget plot]
table {%
-2.75 1.75
-2.75 1.07191780821918
-2.82848837209302 0.997945205479452
-2.9593023255814 0.997945205479452
-2.9593023255814 1.75
-2.9593023255814 0.923972602739726
-3.03779069767442 0.85
-5.99418604651163 0.874657534246575
-5.83720930232558 0.726712328767123
-5.99418604651163 0.874657534246575
-6.75290697674419 0.899315068493151
-7.09302325581395 1.02260273972603
-7.25 1.63904109589041
-7.25 1.75
};
\addplot [TNOblue, forget plot]
table {%
-2.75 1.75
-2.75 2.42808219178082
-2.82848837209302 2.50205479452055
-2.9593023255814 2.50205479452055
-2.9593023255814 1.75
-2.9593023255814 2.57602739726027
-3.03779069767442 2.65
-5.99418604651163 2.62534246575343
-5.83720930232558 2.77328767123288
-5.99418604651163 2.62534246575343
-6.75290697674419 2.60068493150685
-7.09302325581395 2.47739726027397
-7.25 1.86095890410959
-7.25 1.75
};
\addplot [TNOblue, forget plot]
table {%
-4.65988372093023 1.75
-4.65988372093023 0.997945205479452
-4.8953488372093 1.09657534246575
-4.8953488372093 1.75
};
\addplot [TNOblue, forget plot]
table {%
-4.65988372093023 1.75
-4.65988372093023 2.50205479452055
-4.8953488372093 2.40342465753425
-4.8953488372093 1.75
};
\addplot [TNOblue, forget plot]
table {%
-5.83720930232558 1.75
-5.83720930232558 1.44178082191781
-5.75872093023256 1.07191780821918
-6.28197674418605 0.973287671232877
-6.36046511627907 1.46643835616438
-6.36046511627907 1.75
};
\addplot [TNOblue, forget plot]
table {%
-5.83720930232558 1.75
-5.83720930232558 2.05821917808219
-5.75872093023256 2.42808219178082
-6.28197674418605 2.52671232876712
-6.36046511627907 2.03356164383562
-6.36046511627907 1.75
};
\addplot [TNOblue, forget plot]
table {%
-3.11627906976744 1.75
-3.11627906976744 1.02260273972603
-4.52906976744186 1.02260273972603
};
\addplot [TNOblue, forget plot]
table {%
-3.11627906976744 1.75
-3.11627906976744 2.47739726027397
-4.52906976744186 2.47739726027397
};
\addplot [semithick, TNOred, dashed, forget plot]
table {%
2.75 1.75
2.67171717171717 1.74996447297214
2.59343434343434 1.74972010306519
2.51515151515152 1.74906980619819
2.43686868686869 1.74782908517264
2.35858585858586 1.74582576468553
2.28030303030303 1.74289972634234
2.2020202020202 1.73890264367003
2.12373737373737 1.73369771713003
2.04545454545455 1.72715940913127
1.96717171717172 1.71917317904316
1.88888888888889 1.70963521820861
1.81060606060606 1.69845218495698
1.73232323232323 1.68554093961714
1.6540404040404 1.67082827953046
1.57575757575758 1.65425067406376
1.49747474747475 1.63575399962237
1.41919191919192 1.6152932746631
1.34090909090909 1.59283239470727
1.26262626262626 1.56834386735364
1.18434343434343 1.54180854729151
1.10606060606061 1.51321537131363
1.02777777777778 1.48256109332927
0.949494949494949 1.44985001937716
0.871212121212121 1.41509374263854
0.792929292929293 1.37831087845013
0.714646464646465 1.33952679931715
0.636363636363636 1.2987733699263
0.558080808080808 1.25608868215878
0.47979797979798 1.21151679010327
0.401515151515151 1.16510744506897
0.323232323232323 1.11691583059853
0.244949494949495 1.06700229748114
0.166666666666667 1.01543209876543
0.0883838383838382 0.962275124772575
0.0101010101010099 0.907605638109211
-0.0681818181818183 0.85150200868048
-0.146464646464647 0.794046448703017
-0.224747474747475 0.735324747717952
-0.303030303030303 0.67542600760391
-0.381313131313131 0.614442377590011
-0.45959595959596 0.552468789268875
-0.537878787878788 0.489602691609613
-0.616161616161616 0.425943785970835
-0.694444444444445 0.361593761113651
-0.772727272727273 0.296656028214665
-0.851010101010101 0.231235455878979
-0.92929292929293 0.165438105153197
-1.00757575757576 0.0993709645384162
-1.08585858585859 0.033141685003238
-1.16414141414141 -0.0331416850032387
-1.24242424242424 -0.0993709645384178
-1.32070707070707 -0.165438105153197
-1.3989898989899 -0.23123545587898
-1.47727272727273 -0.296656028214666
-1.55555555555556 -0.361593761113652
-1.63383838383838 -0.425943785970836
-1.71212121212121 -0.489602691609613
-1.79040404040404 -0.552468789268874
-1.86868686868687 -0.614442377590012
-1.9469696969697 -0.67542600760391
-2.02525252525253 -0.735324747717952
-2.10353535353535 -0.794046448703018
-2.18181818181818 -0.851502008680481
-2.26010101010101 -0.907605638109211
-2.33838383838384 -0.962275124772574
-2.41666666666667 -1.01543209876543
-2.49494949494949 -1.06700229748113
-2.57323232323232 -1.11691583059853
-2.65151515151515 -1.16510744506897
-2.72979797979798 -1.21151679010327
-2.80808080808081 -1.25608868215878
-2.88636363636364 -1.2987733699263
-2.96464646464647 -1.33952679931715
-3.04292929292929 -1.37831087845013
-3.12121212121212 -1.41509374263854
-3.19949494949495 -1.44985001937716
-3.27777777777778 -1.48256109332927
-3.35606060606061 -1.51321537131363
-3.43434343434343 -1.54180854729151
-3.51262626262626 -1.56834386735365
-3.59090909090909 -1.59283239470727
-3.66919191919192 -1.6152932746631
-3.74747474747475 -1.63575399962237
-3.82575757575758 -1.65425067406375
-3.9040404040404 -1.67082827953045
-3.98232323232323 -1.68554093961715
-4.06060606060606 -1.69845218495698
-4.13888888888889 -1.7096352182086
-4.21717171717172 -1.71917317904316
-4.29545454545455 -1.72715940913127
-4.37373737373737 -1.73369771713003
-4.4520202020202 -1.73890264367003
-4.53030303030303 -1.74289972634234
-4.60858585858586 -1.74582576468553
-4.68686868686869 -1.74782908517264
-4.76515151515152 -1.7490698061982
-4.84343434343434 -1.74972010306519
-4.92171717171717 -1.74996447297215
-5 -1.75
-11 -1.75
};
\addplot [semithick, TNOred, forget plot]
table {%
-10.25 -1
-11 -1.75
-10.25 -2.5
};
\addplot [semithick, TNOblue, dashed, forget plot]
table {%
-7.25 1.75
-9.25 1.75
};
\addplot [semithick, TNOblue, forget plot]
table {%
-8.5 2.5
-9.25 1.75
-8.5 1
};
\end{axis}

\end{tikzpicture}

%% file: figures/events.tikz

\definecolor{TNOred}{RGB}{204,0,0}
\sffamily

\begin{tikzpicture}

\begin{axis}[
xlabel={$t$ [s]},
xmin=-7, xmax=31,
ymin=0, ymax=6,
width=\figurewidth,
height=\figureheight,
tick align=outside,
tick pos=left,
x grid style={white!69.019607843137251!black},
clip marker paths,
ytick=\empty,
xtick={0,5,10,15,20,25,30,35,40},
y axis line style={draw opacity=0}
]
\path [draw=white!80.0!black, fill=white!80.0!black] (axis cs:0,0)
--(axis cs:42.325,0)
--(axis cs:42.325,1)
--(axis cs:0,1)
--cycle;

\path [draw=white!70.0!black, fill=white!70.0!black] (axis cs:0,1)
--(axis cs:42.325,1)
--(axis cs:42.325,2)
--(axis cs:0,2)
--cycle;

\path [draw=white!80.0!black, fill=white!80.0!black] (axis cs:0,2)
--(axis cs:17.7200000286102,2)
--(axis cs:17.7200000286102,3)
--(axis cs:0,3)
--cycle;

\path [draw=white!70.0!black, fill=white!70.0!black] (axis cs:0,3)
--(axis cs:17.7200000286102,3)
--(axis cs:17.7200000286102,4)
--(axis cs:0,4)
--cycle;

\path [draw=white!80.0!black, fill=white!80.0!black] (axis cs:0.800000190734863,4)
--(axis cs:34.1600000858307,4)
--(axis cs:34.1600000858307,5)
--(axis cs:0.800000190734863,5)
--cycle;

\path [draw=white!70.0!black, fill=white!70.0!black] (axis cs:0.800000190734863,5)
--(axis cs:34.1600000858307,5)
--(axis cs:34.1600000858307,6)
--(axis cs:0.800000190734863,6)
--cycle;

\addplot [line width=2.0pt, TNOred, forget plot]
table {%
17 -0.5
17 6
};
\addplot [semithick, black, forget plot]
table {%
0 0
0 1
};
\addplot [semithick, black, forget plot]
table {%
9 0
9 1
};
\addplot [semithick, black, forget plot]
table {%
11 0
11 1
};
\addplot [semithick, black, forget plot]
table {%
18 0
18 1
};
\addplot [semithick, black, forget plot]
table {%
23 0
23 1
};
\addplot [semithick, black, forget plot]
table {%
0 1
0 2
};
\addplot [semithick, black, forget plot]
table {%
17 1
17 2
};
\addplot [semithick, black, forget plot]
table {%
23 1
23 2
};
\addplot [semithick, black, forget plot]
table {%
31.25 1
31.25 2
};
\addplot [semithick, black, forget plot]
table {%
36.25 1
36.25 2
};
\addplot [semithick, black, forget plot]
table {%
42.325 0
42.325 2
};
\addplot [semithick, black, forget plot]
table {%
0 2
0 3
};
\addplot [semithick, black, forget plot]
table {%
0 3
0 4
};
\addplot [semithick, black, forget plot]
table {%
17.7200000286102 2
17.7200000286102 3
};
\addplot [semithick, black, forget plot]
table {%
17.7200000286102 3
17.7200000286102 4
};
\addplot [semithick, black, forget plot]
table {%
13.2000000476837 2
13.2000000476837 3
};
\addplot [semithick, black, forget plot]
table {%
9.20000004768372 3
9.20000004768372 4
};
\addplot [semithick, black, forget plot]
table {%
11.8000001907349 3
11.8000001907349 4
};
\addplot [semithick, black, forget plot]
table {%
0.800000190734863 4
0.800000190734863 5
};
\addplot [semithick, black, forget plot]
table {%
0.800000190734863 5
0.800000190734863 6
};
\addplot [semithick, black, forget plot]
table {%
34.1600000858307 4
34.1600000858307 5
};
\addplot [semithick, black, forget plot]
table {%
34.1600000858307 5
34.1600000858307 6
};
\addplot [semithick, black, forget plot]
table {%
0 1
42.325 1
};
\addplot [semithick, black, forget plot]
table {%
0 2
42.325 2
};
\addplot [semithick, black, forget plot]
table {%
0 3
42.325 3
};
\addplot [semithick, black, forget plot]
table {%
0 4
42.325 4
};
\addplot [semithick, black, forget plot]
table {%
0 5
42.325 5
};
\node at (axis cs:4.4875,0.5)[
  scale=0.75,
  text=black,
  rotate=0.0
]{ Accelerating};
\node at (axis cs:9.9875,0.5)[
  scale=0.75,
  text=black,
  rotate=0.0,
  align=center
]{ Bra-\\
king};
\node at (axis cs:14.4875,0.5)[
  scale=0.75,
  text=black,
  rotate=0.0
]{ Cruising};
\node at (axis cs:20.4875,0.5)[
  scale=0.75,
  text=black,
  rotate=0.0
]{ Accelerating};
\node at (axis cs:27,0.5)[
  scale=0.75,
  text=black,
  rotate=0.0
]{ Cruising};
\node at (axis cs:8.4875,1.5)[
  scale=0.75,
  text=black,
  rotate=0.0
]{ Straight};
\node at (axis cs:19.9875,1.5)[
  scale=0.75,
  text=black,
  rotate=0.0,
  align=center
]{ Lane\\
Change};
\node at (axis cs:27,1.5)[
  scale=0.75,
  text=black,
  rotate=0.0
]{ Straight};
\node at (axis cs:6.60000002384186,2.5)[
  scale=0.75,
  text=black,
  rotate=0.0
]{ Accelerating};
\node at (axis cs:15.460000038147,2.5)[
  scale=0.75,
  text=black,
  rotate=0.0
]{ Cruising};
\node at (axis cs:4.60000002384186,3.5)[
  scale=0.75,
  text=black,
  rotate=0.0
]{ Straight};
\node at (axis cs:10.5000001192093,3.5)[
  scale=0.75,
  text=black,
  rotate=0.0,
  align=center
]{ Lane\\
Change};
\node at (axis cs:14.7600001096725,3.5)[
  scale=0.75,
  text=black,
  rotate=0.0
]{ Straight};
\node at (axis cs:17.4800001382828,4.5)[
  scale=0.75,
  text=black,
  rotate=0.0
]{ Cruising};
\node at (axis cs:17.4800001382828,5.5)[
  scale=0.75,
  text=black,
  rotate=0.0
]{ Straight};
\node at (axis cs:-3.5,0.5)[
  scale=0.75,
  text=black,
  rotate=0.0,
  align=center
]{ Ego vehicle,\\
Longitudinal state};
\node at (axis cs:-3.5,1.5)[
  scale=0.75,
  text=black,
  rotate=0.0,
  align=center
]{ Ego vehicle,\\
Lateral state};
\node at (axis cs:-3.5,2.5)[
  scale=0.75,
  text=black,
  rotate=0.0,
  align=center
]{ Station wagon,\\
Longitudinal state};
\node at (axis cs:-3.5,3.5)[
  scale=0.75,
  text=black,
  rotate=0.0,
  align=center
]{ Station wagon,\\
Lateral state};
\node at (axis cs:-3.5,4.5)[
  scale=0.75,
  text=black,
  rotate=0.0,
  align=center
]{ Pickup truck,\\
Longitudinal state};
\node at (axis cs:-3.5,5.5)[
  scale=0.75,
  text=black,
  rotate=0.0,
  align=center
]{ Pickup truck,\\
Lateral state};
\end{axis}

\end{tikzpicture}

%% file: sections/conclusion.tex
\section{Conclusion}
\label{sec:conclusion}


The need for automated vehicles (AVs) in a dense city such as Singapore will require to have an assessment methodology for road approval authorities in place. Therefore, we proposed an assessment methodology for the safety assessment of AVs that includes functional safety, safety of the intended functionality, and behavioral safety. This, ultimately, results in a qualitative and quantitative assessment of the safety of an AV when operating in real-world traffic.

The proposed methodology adopts a scenario-based assessment, using real-world data to employ a scenario database with scenarios that an AV might encounter when operating in real life. Next to a qualitative assessment of the AV, i.e., the design review, a quantitative assessment is performed by using the scenarios for virtual and physical safety validation. By monitoring the deployment of AVs, new data is acquired that can be used for continuous extension and improvement of the scenario database.

Ultimately, the proposed methodology provides authorities with a formal road approval procedure for AVs. In particular, the proposed methodology will be used to support the Land Transport Authority from Singapore for road approval of AVs.

%% file: sections/acknowledgement.tex
\section*{Acknowledgment}

The research leading to this paper has been realized with the Centre of Excellence for Testing and Research of Autonomous Vehicles at NTU (CETRAN), Singapore. 

%% file: its_ap_2018.bbl
\begin{thebibliography}{10}

\bibitem{Madni2018}
Madni, A.M.:
\newblock Autonomous system-of-systems.
\newblock In: Transdisciplinary Systems Engineering.
\newblock Springer (2018)  161--186

\bibitem{Bimbraw2015}
Bimbraw, K.:
\newblock Autonomous cars: Past, present and future a review of the
  developments in the last century, the present scenario and the expected
  future of autonomous vehicle technology.
\newblock In: 12th International Conference on Informatics in Control,
  Automation and Robotics (ICINCO). Volume~1. (2015)  191--198

\bibitem{Spieser2014}
Spieser, K., Treleaven, K., Zhang, R., Frazzoli, E., Morton, D., Pavone, M.:
\newblock Toward a systematic approach to the design and evaluation of
  automated mobility-on-demand systems: A case study in singapore.
\newblock In: Road Vehicle Automation.
\newblock Springer (2014)  229--245

\bibitem{Bengler2014}
Bengler, K., Dietmayer, K., Farber, B., Maurer, M., Stiller, C., Winner, H.:
\newblock Three decades of driver assistance systems: Review and future
  perspectives.
\newblock IEEE Intelligent Transportation Systems Magazine \textbf{6}(4) (2014)
   6--22

\bibitem{Stellet2015}
Stellet, J.E., Zofka, M.R., Schumacher, J., Schamm, T., Niewels, F.,
  Z\"{o}llner, J.M.:
\newblock Testing of advanced driver assistance towards automated driving: A
  survey and taxonomy on existing approaches and open questions.
\newblock In: IEEE 18th International Conference on Intelligent Transportation
  Systems. (2015)  1455--1462

\bibitem{Helmer2017}
Helmer, T., Kompa{\ss}, K., Wang, L., K{\"u}hbeck, T., Kates, R.
\newblock In: Safety Performance Assessment of Assisted and Automated Driving
  in Traffic: Simulation as Knowledge Synthesis. Springer International
  Publishing (2017)  473--494

\bibitem{Putz2017}
P\"{u}tz, A., Zlocki, A., Bock, J., Eckstein, L.:
\newblock System validation of highly automated vehicles with a database of
  relevant traffic scenarios.
\newblock In: 12th ITS European Congress. (2017)

\bibitem{Wachenfeld2016}
Wachenfeld, W., Winner, H.:
\newblock The release of autonomous vehicles.
\newblock In: Autonomous Driving.
\newblock Springer (2016)  425--449

\bibitem{ISO26262}
{ISO/DIS 26262-1:2011}:
\newblock Road vehicles -- {F}unctional safety -- {P}art 1: {V}ocabulary.
\newblock Technical report, International Organization for Standardization,
  Geneva, Switzerland (2011)

\bibitem{Response2006}
Knapp, A., Neumann, M., Brockmann, M., Walz, R., Winkle, T.:
\newblock Code of practice for the design and evaluation of {ADAS}.
\newblock RESPOSNE III: a PReVENT Project (2009)

\bibitem{ISO21448}
{ISO/WD PAS 21448}:
\newblock Road vehicles -- {S}afety of the intended functionality.
\newblock Technical report, International Organization for Standardization,
  Geneva, Switzerland (Under development)

\bibitem{Waymo2017}
:
\newblock On the road to fully self-driving: {W}aymo safety report (2017)

\bibitem{Gelder2017}
de~Gelder, E., Paardekooper, J.P.:
\newblock Assessment of automated driving systems using real-life scenarios.
\newblock In: Proceedings of the IEEE Intelligent Vehicles Symposium. (2017)
  589--594

\bibitem{Dubey2017}
Dubey, A.:
\newblock The challenges to achieve functional safety for deep learning
  algorithm in automotive.
\newblock Symposium Operational Safe Systems for Automated Driving, Berlin,
  Germany, Sep. 19--20, 2017

\bibitem{Fischer2017}
Fischer, A.:
\newblock {OEM} strategy: {F}rom fail-safe to fail-operational depending on
  {SAE} automation level.
\newblock Symposium Operational Safe Systems for Automated Driving, Berlin,
  Germany, Sep. 19--20, 2017

\bibitem{SAEJ3016}
:
\newblock Taxonomy and definitions for terms related to driving automation
  systems for on-road motor vehicles.
\newblock SAE standard J3016 (September 2016)

\bibitem{Geyer2014}
Geyer, S., Baltzer, M., Franz, B., Hakuli, S., Kauer, M., Kienle, M., Meier,
  S., Weissgerber, T., Bengler, K., Bruder, R., Flemisch, F., Winner, H.:
\newblock Concept and development of a unified ontology for generating test and
  use-case catalogues for assisted and automated vehicle guidance.
\newblock IET Intelligent Transport Systems \textbf{8}(3) (2014)  183--189

\bibitem{Ulbrich2015}
Ulbrich, S., Menzel, T., Reschka, A., Schuldt, F., Maurer, M.:
\newblock Defining and substantiating the terms scene, situation, and scenario
  for automated driving.
\newblock In: 2015 IEEE 18th International Conference on Intelligent
  Transportation Systems. (2015)  982--988

\bibitem{Elrofai2016}
Elrofai, H., Worm, D., Op~den Camp, O.
\newblock In: Scenario identification for validation of automated driving
  functions. Springer International Publishing (2016)  153--163

\bibitem{Gelder2018}
de~Gelder, E., Paardekooper, J.P., Ploeg, J., Elrofai, H., Op~den Camp, O.,
  Khabbaz~Saberi, A., De~Schutter, B.:
\newblock Ontology of scenarios for the assessment of automated vehicles.
\newblock In preparation.

\bibitem{Mahmassani2012}
Mahmassani, H., Mudge, R., Hou, T., Kim, J.:
\newblock Use of mobile data for weather-responsive traffic management models.
\newblock Technical report, U.S. Department of Transportation (2012)

\bibitem{Bonnin2014}
Bonnin, S., Weisswange, T.H., Kummert, F., Schmuedderich, J.:
\newblock General behavior prediction by a combination of scenario-specific
  models.
\newblock IEEE Transactions on Intelligent Transportation Systems
  \textbf{15}(4) (2014)  1478--1488

\end{thebibliography}
